\documentclass[preprint,12pt]{elsarticle}

\usepackage[utf8]{inputenc}
\usepackage[T1]{fontenc}
\usepackage{amsmath,amssymb}
\usepackage{graphicx}
\usepackage{booktabs}
\usepackage{multirow}
\usepackage{hyperref}
\usepackage{xcolor}
\usepackage{array}
\usepackage{float}
\usepackage{caption}
\usepackage{geometry}
\journal{arXiv preprint}

\begin{document}

\begin{frontmatter}

\title{CLIP Architecture for Abdominal CT Image--Text Alignment and Zero-Shot Learning: Investigating Batch Composition and Data Scaling}

\author[pgimer]{Shivika}
\author[pgimer]{Kartik Bose}
\author[pgimer]{Pankaj Gupta\corref{cor1}}
\cortext[cor1]{Corresponding author. Email: Pankajgupta959@gmail.com. Tel: 0172-2756602}
\address[pgimer]{Department of Radiodiagnosis and Imaging, PGIMER, Chandigarh, India-160012}

\begin{abstract}
Vision-language models trained with contrastive learning on paired medical images and reports show strong zero-shot diagnostic capabilities, yet the effect of training batch composition on learned representations remains unexplored for 3D medical imaging. We reproduce Merlin, a dual-encoder model that aligns 3D abdominal CT volumes with radiology reports using symmetric InfoNCE loss, achieving a zero-shot macro F1 of 74.45\% across 30 findings (original: 73.00\%). We then investigate two axes of variation. First, we control the normal-to-abnormal ratio within training batches at 25:75, 50:50, and 75:25 using section-level balanced sampling on the full dataset. All three configurations underperform the unbalanced baseline by 2.4 to 2.8 points, with 75:25 achieving the best result (72.02\%) among balanced variants. Second, we conduct data scaling ablations on a 4,362-study subset, training with 20\%, 40\%, and 100\% of the data. Performance scales sub-linearly from 65.26\% to 71.88\%, with individual findings varying dramatically in data sensitivity. Enforcing 50:50 balanced sampling on the same subset further degrades performance to 68.01\%, confirming that explicit class balancing hurts regardless of dataset or balancing granularity. Our results indicate that the stochastic diversity of random sampling, combined with Merlin's alternating batching over anatomical subsections, provides more effective regularization than engineered class ratios at the small batch sizes required by 3D medical volumes.
\end{abstract}


\end{frontmatter}

% ══════════════════════════════════════════════════════════════
\section{Introduction}
\label{sec:introduction}
% ══════════════════════════════════════════════════════════════

Vision-language models (VLMs) learn shared representations from paired image and text data through contrastive learning. Models such as CLIP~\cite{radford2021clip} and ALIGN~\cite{jia2021align} demonstrated that aligning visual and textual embeddings in a shared space enables powerful zero-shot transfer to downstream tasks. This paradigm has since been adapted to medical imaging, where paired images and clinical reports provide a natural source of supervision~\cite{zhang2022convirt,huang2021gloria,boecking2022biovil}.

Abdominal computed tomography (CT) presents unique challenges for VLMs. Volumes are three-dimensional, reports are long and structured around multiple anatomical regions, and pathological findings are sparse relative to normal observations. Merlin~\cite{blankemeier2024merlin} addressed these challenges with a dual-encoder architecture that pairs a 3D ResNet152 I3D vision encoder with a Clinical Longformer text encoder, trained using symmetric InfoNCE loss on 15,331 CT-report pairs. The model achieved strong zero-shot classification across 30 abdominal findings without any task-specific fine-tuning.

A notable design choice in Merlin is its alternating batching strategy. On even-numbered training steps, the model sees full radiology reports. On odd-numbered steps, the model cycles through 12 anatomical subsections, presenting only the text for a single body region. This approach exposes the model to both holistic and region-specific associations. However, the authors did not investigate how the composition of training batches, specifically the ratio of normal to abnormal studies, affects learned representations.

Batch composition is a known concern in contrastive learning. Class-imbalanced batches can bias the learned embedding space~\cite{zhu2022balanced,cui2021parametric}, and the selection of hard negatives has been shown to significantly affect representation quality~\cite{robinson2021hard,kalantidis2020hard}. In medical datasets, where normal cases often dominate, the ratio of normal to abnormal samples within each batch may influence how effectively the model learns to distinguish pathological patterns. Separately, the relationship between training set size and contrastive learning performance has been studied in natural image domains~\cite{chen2020simclr,he2020moco,yeh2022decoupled}, but remains underexplored for medical VLMs.

This work makes three contributions:
\begin{enumerate}
    \item We reproduce the Merlin framework from scratch, achieving a zero-shot macro F1 of 74.45\% compared to the original 73.00\%.
    \item We investigate the effect of batch-level normal-to-abnormal ratio on zero-shot performance, training models at 25:75, 50:50, and 75:25 ratios using section-level balanced sampling on the full training set.
    \item We conduct data scaling ablations on a curated normal/abnormal-labeled subset, training with 20\%, 40\%, and 100\% of the data to quantify the relationship between dataset size and downstream performance.
\end{enumerate}
Our results show that while increasing the proportion of normal studies yields modest gains within the balanced-sampling family, the original alternating strategy without explicit class balancing remains superior.

% ══════════════════════════════════════════════════════════════
\section{Related Work}
\label{sec:related}
% ══════════════════════════════════════════════════════════════

\subsection{Vision-Language Models in Medical Imaging}

Contrastive learning between medical images and clinical text has emerged as a dominant paradigm for self-supervised representation learning in radiology. ConVIRT~\cite{zhang2022convirt} first demonstrated that training a dual-encoder model on paired chest X-rays and reports yields representations that transfer well to downstream classification tasks. GLoRIA~\cite{huang2021gloria} extended this idea with attention-based local alignment between image regions and text tokens. BioViL~\cite{boecking2022biovil} combined masked language modeling with image-text contrastive learning for chest X-ray interpretation. CheXzero~\cite{tiu2022chexzero} showed that a CLIP-style model trained on chest X-rays achieves expert-level zero-shot pathology detection. These works established that medical VLMs can learn clinically meaningful representations without explicit labels, but they focused exclusively on 2D imaging modalities.

\subsection{3D Medical Vision-Language Models}

Extending VLMs to 3D volumes introduces additional challenges in both computational cost and text alignment. CT-CLIP~\cite{hamamci2024ctclip} adapted the CLIP framework to volumetric chest CT scans using a video-encoder-style architecture. Merlin~\cite{blankemeier2024merlin} addressed abdominal CT specifically, using an inflated 3D ResNet152 for the vision encoder and a Clinical Longformer for the text encoder. Merlin introduced an alternating batching strategy that cycles between full reports and individual anatomical subsections, enabling the model to learn both global and region-specific associations. The authors reported a macro F1 of 73.5\% across 30 zero-shot findings classifications, establishing a strong baseline for abdominal CT VLMs.

\subsection{Batch Composition and Class Imbalance in Contrastive Learning}

The composition of training batches affects contrastive learning in several ways. Zhu et al.~\cite{zhu2022balanced} showed that class-imbalanced data leads to biased representations in the contrastive embedding space, where minority-class samples are pushed toward majority-class clusters. Cui et al.~\cite{cui2021parametric} proposed parametric contrastive learning to mitigate class imbalance by adjusting the contrastive objective based on class frequencies. Hard negative mining strategies~\cite{robinson2021hard,kalantidis2020hard} demonstrated that the selection of informative negatives within each batch is critical for learning discriminative representations. These findings suggest that the normal-to-abnormal ratio within training batches may influence how effectively medical VLMs distinguish pathological from normal findings.

\subsection{Batch Size and Data Scaling in Contrastive Learning}

Contrastive learning performance is sensitive to both batch size and dataset scale. Chen et al.~\cite{chen2020simclr} showed that SimCLR benefits substantially from larger batch sizes, as more negatives per positive improve the quality of the learned representations. He et al.~\cite{he2020moco} decoupled batch size from the number of negatives through a momentum-updated queue in MoCo. Yeh et al.~\cite{yeh2022decoupled} provided theoretical analysis showing that larger batches reduce the variance of the InfoNCE gradient estimator. On the data scaling front, Cherti et al.~\cite{cherti2023scaling} demonstrated log-linear scaling between dataset size and downstream performance for CLIP models trained on web-crawled data. Whether similar scaling relationships hold for medical VLMs trained on comparatively small, curated datasets remains an open question.

\subsection{Positioning of Our Work}

Prior work on medical VLMs has not systematically studied how batch-level class composition affects learned representations. Studies on class imbalance in contrastive learning~\cite{zhu2022balanced,cui2021parametric} have focused on the overall dataset distribution rather than per-batch control. Our work fills this gap by explicitly controlling the normal-to-abnormal ratio at the batch level during contrastive pretraining of a 3D medical VLM. We also provide data scaling ablations specific to the medical domain, complementing the large-scale scaling analyses that have been conducted for natural image CLIP models~\cite{cherti2023scaling}.

% ══════════════════════════════════════════════════════════════
\section{Methods}
\label{sec:methods}
% ══════════════════════════════════════════════════════════════

\subsection{Dataset and Preprocessing}

We use a dataset of 25,494 abdominopelvic CT studies from Stanford Hospital Emergency Department, each paired with a radiology report. Each report is parsed into 12 anatomical subsections: lower thorax, liver and biliary system, gallbladder, spleen, pancreas, adrenal glands, kidneys and ureters, gastrointestinal tract, peritoneal cavity, pelvic organs, vasculature and lymph nodes, and musculoskeletal system. The dataset is split at the patient level into 15,314 training, 5,055 validation, and 5,125 test studies (approximately 60/20/20).

CT volumes are preprocessed to a fixed spatial resolution of $224 \times 224 \times 160$ voxels and stored as NumPy arrays. Reports are tokenized using the Clinical Longformer tokenizer with a maximum sequence length of 4,096 tokens. For each study, we store both the full concatenated report and the 12 individual subsection texts.

For data scaling experiments, we construct a Normal/Abnormal (NAB) subset of 4,362 studies with binary case-level labels. A study is labeled normal only if all 12 anatomical subsections are independently annotated as normal. Similarly, abnormal study had some abnormality in all 12 anatomical subsections. The NAB subset contains 1,982 normal (45.4\%) and 2,380 abnormal (54.6\%) studies, split 80/20 into 3,489 training and 873 validation samples with stratified sampling. We further subsample this set to create 40\% (1,745 studies) and 20\% (872 studies) fractions, preserving the original normal/abnormal ratio.

\subsection{Model Architecture}

The model follows a dual-encoder CLIP-style architecture with separate vision and text encoders projecting into a shared 512-dimensional embedding space.

\textbf{Vision Encoder.} We use a ResNet152~\cite{he2016resnet} inflated to 3D (I3D) as the vision encoder. All 2D convolutions are replaced with 3D convolutions by replicating kernel weights along the depth dimension and normalizing by kernel depth. The stem convolution uses a depth stride of 1 to preserve spatial resolution along the z-axis. The encoder produces a 2,048-dimensional feature vector via global average pooling.

\textbf{Text Encoder.} We use the Clinical Longformer~\cite{li2022clinical}, a domain-adapted variant of the Longformer architecture pretrained on clinical notes from MIMIC-III. The model supports sequences up to 4,096 tokens using a combination of local sliding-window attention and global attention on the [CLS] token. The [CLS] token representation (768-dimensional) serves as the text embedding.

\textbf{Projection Heads.} Each encoder output is passed through a linear projection layer that maps to a 512-dimensional space, followed by L2 normalization. The resulting unit-norm embeddings are used for contrastive loss computation.

\subsection{Training Objective}

We train the model using symmetric InfoNCE loss~\cite{oord2018cpc}. Given a batch of $N$ image-text pairs, we compute the cosine similarity between all image and text embeddings. Let $s(i,j)$ denote the cosine similarity between image embedding $i$ and text embedding $j$, scaled by a learnable temperature parameter $\tau$ (initialized to 0.07). The image-to-text loss is:
\begin{equation}
    \mathcal{L}_{\text{i2t}} = -\frac{1}{N} \sum_{i} \log \frac{\exp(s(i,i)/\tau)}{\sum_{j} \exp(s(i,j)/\tau)}
\end{equation}
The text-to-image loss $\mathcal{L}_{\text{t2i}}$ is defined symmetrically. The total loss is $\mathcal{L} = (\mathcal{L}_{\text{i2t}} + \mathcal{L}_{\text{t2i}}) / 2$.

\subsection{Alternating Batching Strategy}

Following Merlin~\cite{blankemeier2024merlin}, we use an alternating batching strategy during training. On even-numbered training steps, the text input consists of the full concatenated radiology report. On odd-numbered steps, we cycle through the 12 anatomical subsections in order, using only the text for the current subsection. This alternation exposes the model to both holistic image-report associations and fine-grained region-specific correspondences within each epoch.

\subsection{Training Configuration}

All experiments share a common training configuration unless otherwise specified. We train with a batch size of 8 using the AdamW optimizer with betas $(0.9, 0.999)$. The learning rate follows a cosine annealing schedule with linear warmup over the first 10\% of training steps. The initial learning rate ramps linearly from $\text{lr}/25$ to the peak learning rate of $1 \times 10^{-5}$, then decays to zero following a cosine curve. Gradient norms are clipped to 1.0. Training uses FP16 mixed precision. The maximum number of training epochs is 300.

\begin{table}[H]
\centering
\caption{Training Hyperparameters.}
\label{tab:hyperparams}
\begin{tabular}{ll}
\toprule
\textbf{Parameter} & \textbf{Value} \\
\midrule
Batch size & 8 \\
Optimizer & AdamW \\
Peak learning rate & $1 \times 10^{-5}$ \\
LR schedule & Cosine annealing with linear warmup \\
Warmup fraction & 10\% of total steps \\
Warmup start LR & lr/25 = $4 \times 10^{-7}$ \\
Temperature $\tau$ & 0.07 (learnable) \\
Gradient clipping & Max norm 1.0 \\
Precision & FP16 mixed precision \\
Max epochs & 300 \\
\bottomrule
\end{tabular}
\end{table}

\subsection{Experiment Family 1: Batch Composition on Full Dataset}

To investigate how the normal-to-abnormal ratio within training batches affects zero-shot performance, we train three models on the full 15,314-study training set using a section-level balanced sampler. This sampler constructs each batch such that the specified ratio of normal to abnormal studies is maintained at the anatomical section level. A study's section is considered normal if its corresponding subsection text indicates no abnormality.

We train models at three ratios (normal:abnormal): 25:75, 50:50, and 75:25. Each model uses the same architecture, hyperparameters, and alternating batching strategy as the reproduction baseline. The only difference is the batch-level class composition enforced by the sampler.

The reproduction baseline uses the default data loading order (standard shuffled sampling without explicit class balancing) on the same 15,314 training studies. All models are evaluated on the same held-out test set of 5,125 studies.

\begin{table}[H]
\centering
\caption{Experiment Family 1: Batch Composition Configurations.}
\label{tab:exp1_config}
\begin{tabular}{lllll}
\toprule
\textbf{Experiment} & \textbf{Normal:Abnormal} & \textbf{Dataset} & \textbf{Sampler} & \textbf{Training Size} \\
\midrule
Baseline (Reproduction) & Natural distribution & Full & Shuffled & 15,314 \\
Ratio 25:75 & 25:75 & Full & SectionBalancedSampler & 15,314 \\
Ratio 50:50 & 50:50 & Full & SectionBalancedSampler & 15,314 \\
Ratio 75:25 & 75:25 & Full & SectionBalancedSampler & 15,314 \\
\bottomrule
\end{tabular}
\end{table}

\subsection{Experiment Family 2: Data Scaling Ablations}

To quantify the effect of training set size on zero-shot classification, we train models on the NAB subset at three data fractions: 100\% (3,489 training studies), 40\% (1,396 studies), and 20\% (697 studies). All three use standard shuffled sampling without class balancing.

We also include a balanced-sampling ablation that enforces a 50:50 normal-to-abnormal ratio at the case level on the full NAB training set using a separate batch sampler (NABBatchSampler). This differs from the section-level balancing in Experiment Family 1 in two ways: it operates at the case level rather than the section level, and it uses the smaller NAB subset rather than the full dataset.

\begin{table}[H]
\centering
\caption{Experiment Family 2: Data Scaling Configurations.}
\label{tab:exp2_config}
\begin{tabular}{llll}
\toprule
\textbf{Data Fraction} & \textbf{Training Studies} & \textbf{Sampler} & \textbf{Normal:Abnormal} \\
\midrule
100\% & 3,489 & Shuffled & Natural (45:55) \\
100\% & 3,489 & NABBatchSampler & 50:50 (enforced) \\
40\% & 1,396 & Shuffled & Natural (45:55) \\
20\% & 697 & Shuffled & Natural (45:55) \\
\bottomrule
\end{tabular}
\end{table}

\subsection{Zero-Shot Evaluation Protocol}

We evaluate all models on zero-shot classification of 30 binary findings. For each finding, the test set is balanced to contain equal numbers of positive and negative cases. For each test study, we compute the mean cosine similarity between the image embedding and a set of positive text prompts describing the finding, and separately between the image embedding and a set of negative text prompts. A study is classified as positive for the finding if the mean positive similarity exceeds the mean negative similarity. We report the per-finding F1 score and the macro F1 averaged across all 30 findings.

% ══════════════════════════════════════════════════════════════
\section{Experiments and Results}
\label{sec:results}
% ══════════════════════════════════════════════════════════════

\subsection{Reproduction of Merlin}

We first verified our reimplementation by training on the full 15,314-study training set with the original alternating batching strategy and no explicit class balancing. Table~\ref{tab:reproduction} compares our reproduction against the original Merlin model.

\begin{table}[H]
\centering
\caption{Reproduction Comparison.}
\label{tab:reproduction}
\begin{tabular}{lcc}
\toprule
\textbf{Model} & \textbf{Macro F1 (\%)} & \textbf{Best Epoch} \\
\midrule
Merlin (original pretrained) & 73.45 & -- \\
Our reproduction & 74.45 & 187 \\
\bottomrule
\end{tabular}
\end{table}

Our reproduction achieves a macro F1 of 74.45\%, exceeding the original by 1.0 percentage point. This confirms the fidelity of our reimplementation and provides a reliable baseline for subsequent experiments.

Table~\ref{tab:perfinding} shows per-finding F1 scores comparing the original Merlin model to our reproduction. Both models achieve high performance on findings with distinctive imaging signatures, such as surgically absent gallbladder (95.01\% vs.\ 89.86\%) and aortic valve calcification (82.52\% vs.\ 90.67\%). The reproduction shows notable improvements on osteopenia ($65.18\% \to 81.41\%$), renal hypodensities ($46.43\% \to 68.40\%$), and pancreatic atrophy ($58.65\% \to 73.99\%$), while the original performs better on splenomegaly (85.71\% vs.\ 73.38\%) and surgically absent gallbladder (95.01\% vs.\ 89.86\%).

\begin{table}[H]
\centering
\caption{Per-Finding F1 Scores: Reproduction vs.\ Original.}
\label{tab:perfinding}
\small
\begin{tabular}{lccc}
\toprule
\textbf{Finding} & \textbf{Original (\%)} & \textbf{Reproduction (\%)} & \textbf{$\Delta$} \\
\midrule
Abdominal aortic aneurysm & 75.68 & 82.93 & +7.25 \\
Anasarca & 88.95 & 79.80 & $-$9.15 \\
Aortic valve calcification & 82.52 & 90.67 & +8.15 \\
Appendicitis & 68.38 & 62.92 & $-$5.46 \\
Ascites & 81.05 & 87.33 & +6.28 \\
Atelectasis & 71.38 & 72.17 & +0.79 \\
Atherosclerosis & 92.17 & 90.09 & $-$2.08 \\
Biliary ductal dilation & 73.87 & 69.23 & $-$4.64 \\
Bowel obstruction & 75.00 & 83.44 & +8.44 \\
Cardiomegaly & 78.95 & 77.90 & $-$1.05 \\
Coronary calcification & 76.67 & 76.34 & $-$0.33 \\
Fracture & 68.85 & 58.73 & $-$10.12 \\
Free air & 76.00 & 77.23 & +1.23 \\
Gallstones & 73.20 & 63.79 & $-$9.41 \\
Hepatic steatosis & 54.29 & 62.75 & +8.46 \\
Hepatomegaly & 74.80 & 80.34 & +5.54 \\
Hiatal hernia & 65.94 & 63.58 & $-$2.36 \\
Hydronephrosis & 67.58 & 67.23 & $-$0.35 \\
Lymphadenopathy & 75.00 & 77.78 & +2.78 \\
Metastatic disease & 73.58 & 70.69 & $-$2.89 \\
Osteopenia & 65.18 & 81.41 & +16.23 \\
Pancreatic atrophy & 58.65 & 73.99 & +15.34 \\
Pleural effusion & 83.19 & 86.18 & +2.99 \\
Prostatomegaly & 73.58 & 65.92 & $-$7.66 \\
Renal cyst & 60.64 & 66.50 & +5.86 \\
Renal hypodensities & 46.43 & 68.40 & +21.97 \\
Splenomegaly & 85.71 & 73.38 & $-$12.33 \\
Submucosal edema & 72.88 & 70.11 & $-$2.77 \\
Surgically absent gallbladder & 95.01 & 89.86 & $-$5.15 \\
Thrombosis & 68.49 & 62.86 & $-$5.63 \\
\midrule
\textbf{Macro F1} & \textbf{73.45} & \textbf{74.45} & \textbf{+1.00} \\
\bottomrule
\end{tabular}
\end{table}

\subsection{Experiment Family 1: Batch Composition}

Table~\ref{tab:batch_results} reports the zero-shot macro F1 for each batch composition experiment at its best-performing checkpoint, compared to the reproduction baseline.

\begin{table}[H]
\centering
\caption{Batch Composition Results: Best Macro F1.}
\label{tab:batch_results}
\begin{tabular}{llcc}
\toprule
\textbf{Experiment} & \textbf{Normal:Abnormal} & \textbf{Best Macro F1 (\%)} & \textbf{Best Epoch} \\
\midrule
Baseline (Reproduction) & Natural (shuffled) & 74.45 & 187 \\
Ratio 75:25 & 75:25 & 72.02 & 15 \\
Ratio 50:50 & 50:50 & 71.97 & 33 \\
Ratio 25:75 & 25:75 & 71.70 & 19 \\
\bottomrule
\end{tabular}
\end{table}

\begin{figure}[H]
\centering
\includegraphics[width=0.85\textwidth]{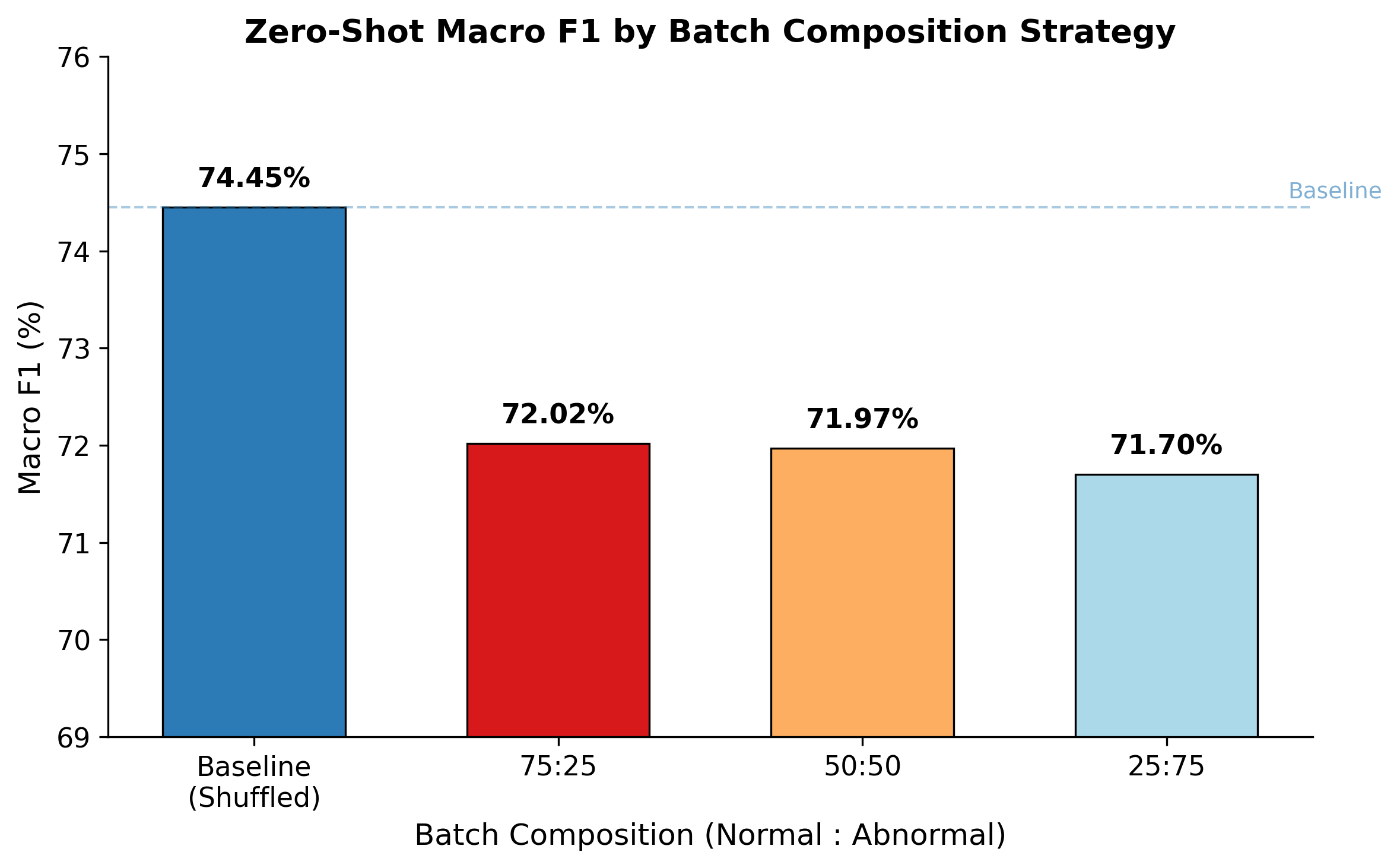}
\caption{Zero-shot macro F1 by batch composition strategy. The baseline with shuffled sampling (no class balancing) outperforms all three balanced-sampling ratios by 2.4--2.8 percentage points. Within the balanced family, higher normal proportion yields marginally better results (75:25 $>$ 50:50 $>$ 25:75).}
\label{fig:batch_composition}
\end{figure}

All three balanced-sampling configurations underperform the reproduction baseline by 2.4 to 2.8 percentage points (Figure~\ref{fig:batch_composition}). Within the balanced-sampling family, there is a consistent trend: increasing the proportion of normal studies in each batch yields modest improvements. The 75:25 ratio achieves the highest F1 (72.02\%), followed by 50:50 (71.97\%) and 25:75 (71.70\%). The differences among the three ratios are small (0.32 points between best and worst), but the ordering is consistent.

\begin{figure}[H]
\centering
\includegraphics[width=\textwidth]{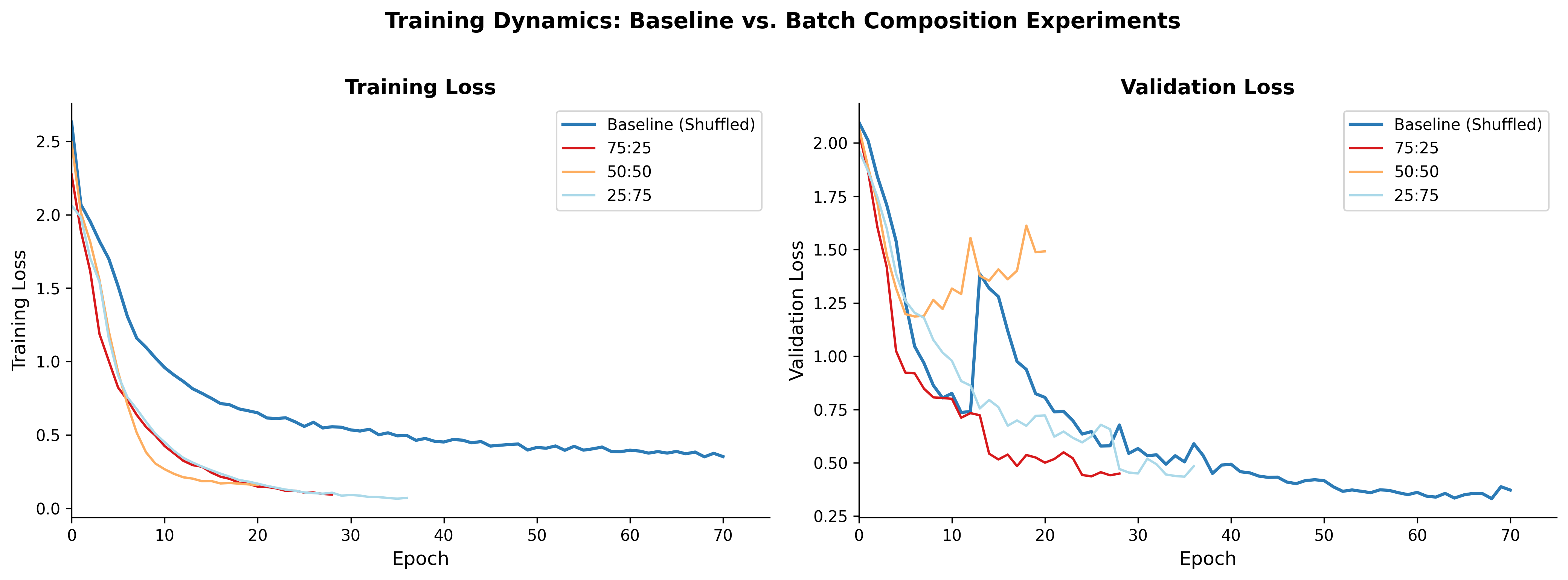}
\caption{Training and validation loss curves for the baseline and batch composition experiments. Left: training loss. Right: validation loss. The balanced-sampling models (75:25, 50:50, 25:75) overfit substantially faster than the baseline, with training loss dropping below 0.1 while validation loss stagnates or rises.}
\label{fig:training_dynamics}
\end{figure}

\textbf{Training Dynamics.} The balanced-sampling models converge faster in terms of training loss but show signs of overfitting (Figure~\ref{fig:training_dynamics}). The 25:75 model reaches a training loss of 0.066 by epoch 35, compared to 0.35 for the baseline at epoch 70, yet its validation loss plateaus at 0.43, well above the baseline's 0.33. The 50:50 model shows the most severe divergence, with validation loss rising from 1.19 (epoch 7) to 1.49 (epoch 20) while training loss continues to decrease. All balanced-sampling models were stopped before 300 epochs due to overfitting or resource constraints.

\subsection{Experiment Family 2: Data Scaling Ablations}

Table~\ref{tab:scaling_results} reports the best zero-shot macro F1 for each data scaling ablation.

\begin{table}[H]
\centering
\caption{Data Scaling Results: Best Macro F1.}
\label{tab:scaling_results}
\begin{tabular}{llccc}
\toprule
\textbf{Ablation} & \textbf{Data Fraction} & \textbf{Training Studies} & \textbf{Best Macro F1 (\%)} & \textbf{Best Epoch} \\
\midrule
100\%, random & 100\% & 3,489 & 71.88 & 25 \\
100\%, balanced (50:50) & 100\% & 3,489 & 68.01 & $\sim$13 \\
40\%, random & 40\% & 1,396 & 67.96 & 90 \\
20\%, random & 20\% & 697 & 65.26 & 30 \\
\bottomrule
\end{tabular}
\end{table}

\begin{figure}[H]
\centering
\includegraphics[width=0.85\textwidth]{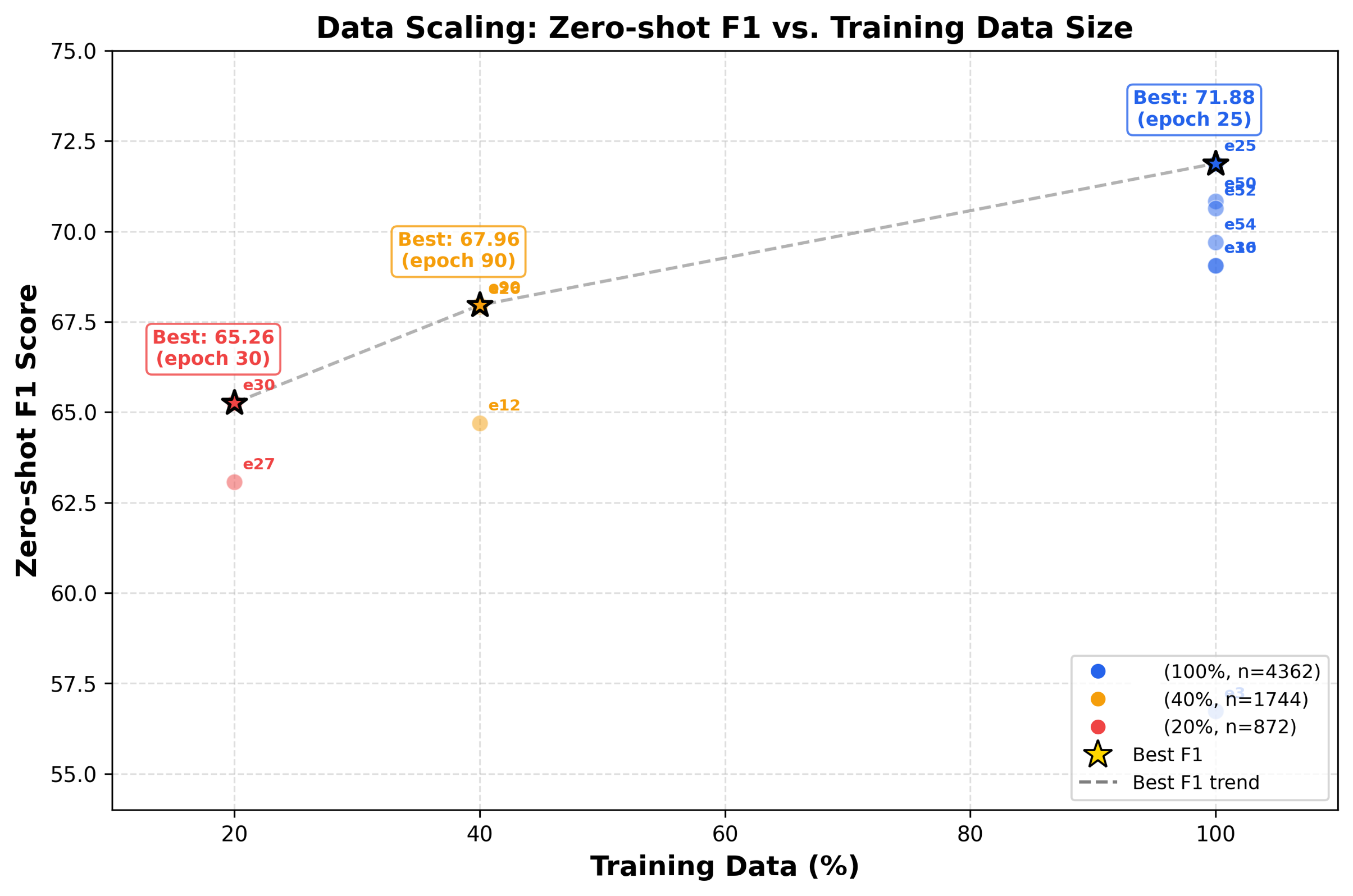}
\caption{Data scaling: zero-shot F1 vs.\ training data size for ablations 100\% NAB, 40\% NAB, and 20\% NAB. Each point represents a checkpoint; stars mark the best F1 for each ablation. The dashed trend line illustrates the sub-linear scaling relationship.}
\label{fig:data_scaling}
\end{figure}

\textbf{Data scaling trend.} Among the randomly-sampled ablations, there is a clear monotonic relationship between training set size and downstream performance (Figure~\ref{fig:data_scaling}). Reducing the dataset from 100\% to 40\% decreases F1 by 3.9 points ($71.88 \to 67.96$), and further reducing to 20\% decreases it by an additional 2.7 points ($67.96 \to 65.26$). The marginal return of additional data diminishes: the first doubling (20\% $\to$ 40\%) yields 2.7 points while the next 2.5$\times$ increase (40\% $\to$ 100\%) yields 3.9 points. This sub-linear scaling is consistent with log-linear relationships observed in larger-scale CLIP training~\cite{cherti2023scaling}.

\textbf{Balanced vs.\ random sampling.} Enforcing 50:50 case-level balancing on the same 3,489 studies achieves an F1 of 68.01\%, underperforming the random (71.88\%) by 3.9 points. This result mirrors the finding from Experiment Family 1 that explicit class balancing degrades performance relative to natural sampling, even when the underlying class distribution is already near-balanced (45:55 in the NAB subset).

\textbf{Training dynamics.} 50:50 case-level balancing shows catastrophic overfitting, with training loss collapsing to near zero (0.002 by epoch 41) while validation loss remains in the 1.5--1.9 range. The other ablations overfit less severely but still shows a large train-validation gap (Figure~\ref{fig:training_dynamics}).

\subsection{Cross-Family Comparison}

Table~\ref{tab:summary} provides an overview of all experiments, enabling comparison across both families.

\begin{table}[H]
\centering
\caption{Summary of All Experiments. NAB: normal--abnormal.}
\label{tab:summary}
\small
\begin{tabular}{llllcc}
\toprule
\textbf{Experiment} & \textbf{Dataset} & \textbf{Training Size} & \textbf{Sampler} & \textbf{Best F1 (\%)} & \textbf{Best Epoch} \\
\midrule
Reproduction baseline & Full & 15,314 & Shuffled & 74.45 & 187 \\
75:25 ratio & Full & 15,314 & SectionBalanced & 72.02 & 15 \\
50:50 ratio & Full & 15,314 & SectionBalanced & 71.97 & 33 \\
25:75 ratio & Full & 15,314 & SectionBalanced & 71.70 & 19 \\
100\%, random & NAB & 3,489 & Shuffled & 71.88 & 25 \\
100\%, balanced & NAB & 3,489 & NABBatchSampler & 68.01 & $\sim$13 \\
40\%, random & NAB & 1,396 & Shuffled & 67.96 & 90 \\
20\%, random & NAB & 697 & Shuffled & 65.26 & 30 \\
\bottomrule
\end{tabular}
\end{table}

Several patterns emerge from this comparison. First, the reproduction baseline trained on the full dataset with simple shuffled sampling achieves the highest F1, outperforming all other configurations. Second, explicit class balancing consistently degrades performance regardless of the balancing granularity (section-level in Family 1 or case-level in Family 2) or the dataset used. Third, 100\% random NAB strategy achieves a macro F1 of 71.88\% using only 3,489 training studies with random sampling, which is within 0.2 points of the best balanced-sampling result on the full 15,314-study dataset (72.02\%). This suggests that dataset composition may matter less than training strategy when the underlying data quality is sufficient.

\subsection{Analysis of Findings by Difficulty}

\begin{figure}[H]
\centering
\includegraphics[width=0.85\textwidth]{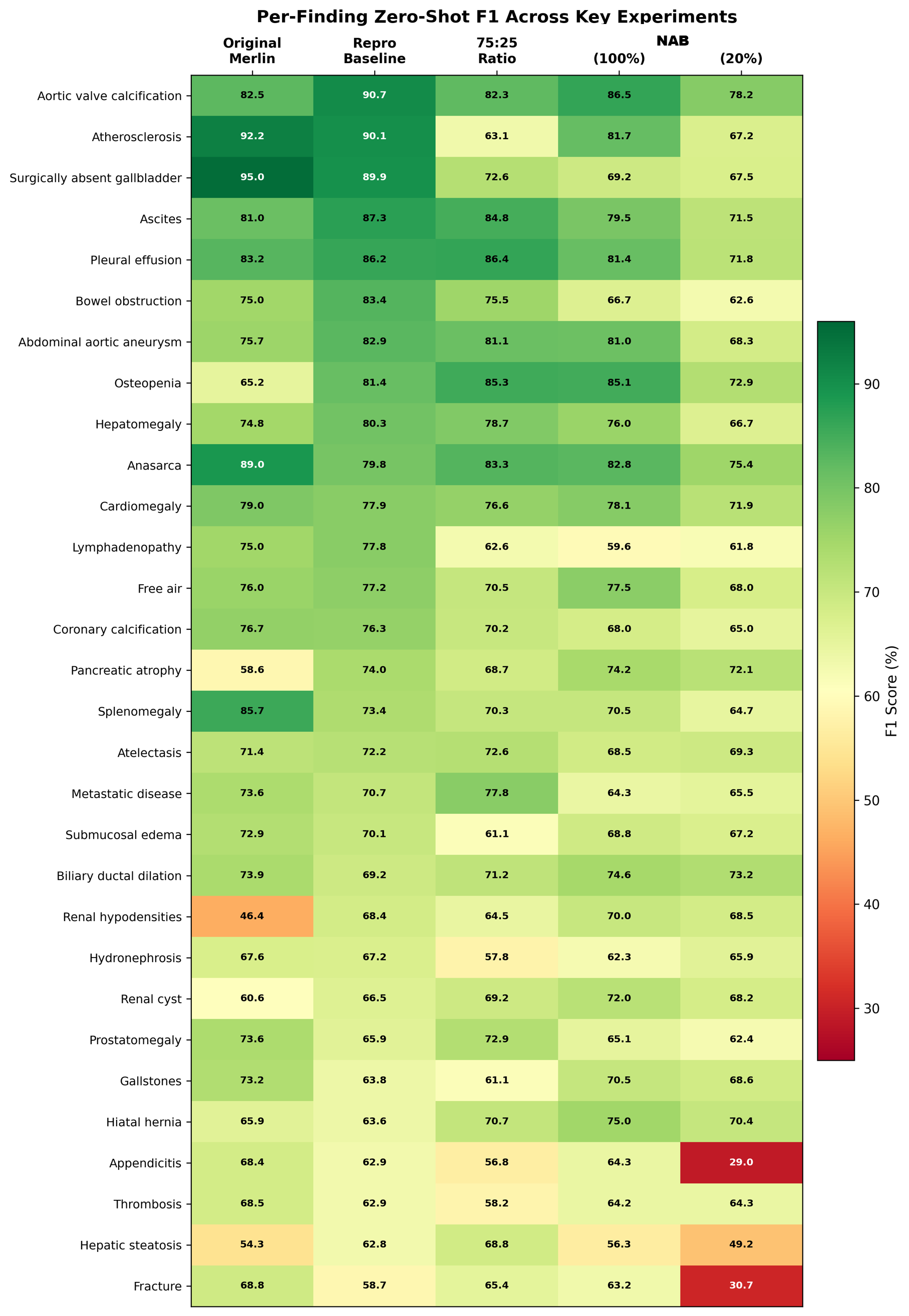}
\caption{Per-finding zero-shot F1 scores across key experiments: original Merlin, our reproduction baseline, 75:25 batch ratio, 100\% NAB data, and 20\% NAB data. Findings are sorted by reproduction baseline F1. Color intensity reflects F1 magnitude, highlighting which findings are most sensitive to training conditions.}
\label{fig:heatmap}
\end{figure}

To better understand which pathologies drive performance differences, we compare per-finding F1 scores across key experiments. Figure~\ref{fig:heatmap} presents a heatmap of F1 scores for all 30 findings across five representative configurations.

Several findings are disproportionately sensitive to data reduction. Appendicitis drops from 64.29\% (100\% random) to 29.03\% (20\% random), a 35-point decline. Fracture drops from 63.19\% to 30.71\%. These are findings with lower prevalence and more variable imaging presentations, making them harder to learn from limited data. Conversely, some findings are relatively robust to data reduction. Pancreatic atrophy shows only a 2-point decline from 100\% NAB to 20\% NAB ($74.18\% \to 72.06\%$). Hiatal hernia drops by less than 5 points ($75.04\% \to 70.41\%$). These findings may have more consistent imaging signatures that can be captured even from small training sets.

% ══════════════════════════════════════════════════════════════
\section{Discussion}
\label{sec:discussion}
% ══════════════════════════════════════════════════════════════

The most consistent finding across both experiment families is that the original Merlin alternating batching strategy, which applies no explicit normal/abnormal balancing, outperforms all configurations that enforce a fixed class ratio. The reproduction baseline achieves a macro F1 of 74.45\%, compared to 72.02\% for the best balanced-sampling model (75:25) and 71.88\% for the best NAB ablation. This gap of 2.4 points persists despite the balanced-sampling models using the same 15,314 training studies and identical hyperparameters.

One explanation is that the alternating strategy provides a form of implicit curriculum. By cycling between full reports and individual anatomical subsections, the model sees each training sample under 13 different textual framings per epoch (one full-report pass and 12 section passes). This variety of text contexts may serve as a regularizer, preventing the model from overfitting to superficial correlations between image features and specific textual patterns. In contrast, the balanced samplers constrain which studies appear in each batch, reducing the effective diversity of image-text pairings the model encounters per epoch.

The training dynamics support this interpretation. The balanced-sampling models consistently overfit faster, with training loss dropping below 0.1 while validation loss stagnates or diverges. The 50:50 model shows the most extreme case, with validation loss rising from epoch 7 onward. The baseline model, by contrast, continues improving its validation loss through epoch 68 and reaches the lowest validation loss of any experiment (0.33). This suggests that the alternating strategy provides better gradient signal for generalization, not merely more training steps.

Within the balanced-sampling family, increasing the proportion of normal studies yields a small but consistent improvement: 75:25 (72.02\%) $>$ 50:50 (71.97\%) $>$ 25:75 (71.70\%). The differences are modest (0.32 points between best and worst), but the ordering is monotonic and holds at multiple checkpoints.

This trend has a plausible mechanistic explanation. In contrastive learning, the InfoNCE loss treats all non-matching pairs within a batch as negatives. When more normal studies are present in a batch, the negative pairs for abnormal samples are more likely to include normals, which are typically easier to distinguish from abnormal cases. This may provide more informative gradient signal for learning abnormal-specific features. Conversely, batches dominated by abnormal studies contain many abnormal-abnormal negative pairs, which may be harder to separate because different abnormal findings can share overlapping imaging features such as mass effect, fluid collections, or organ enlargement.

However, the practical significance of this finding is limited. The 0.32-point range across ratios is small compared to the 2.4-point gap between any balanced-sampling model and the baseline. Practitioners designing batch composition strategies for medical VLMs would benefit more from preserving the natural data distribution and focusing on text augmentation strategies such as alternating batching.

The data scaling ablations reveal a monotonic relationship between training set size and zero-shot performance on the NAB subset: 697 studies yield 65.26\%, 1,396 yield 67.96\%, and 3,489 yield 71.88\%. The marginal gain per additional study decreases with scale, consistent with the log-linear scaling observed by Cherti et al.~\cite{cherti2023scaling} for web-scale CLIP training.

A notable comparison arises between 100\% NAB and the batch composition experiments. 100\% NAB uses only 3,489 training studies with random sampling and achieves 71.88\%, which is within 0.2 points of the 75:25 model trained on 15,314 studies (72.02\%). This comparison is imperfect because batch composition experiments use different datasets (NAB subset vs.\ full training set) and different evaluation conditions. Nevertheless, it suggests that a well-curated smaller dataset with known class labels may approach the performance of a larger dataset with controlled batch composition, provided the sampling strategy does not artificially constrain batch diversity.

The failure of 50:50 balanced (68.01\%) relative to random (71.88\%) on the same 3,489 studies is particularly informative. The NAB subset is already near-balanced at 45:55 normal:abnormal. Enforcing exact 50:50 sampling therefore changes the effective data distribution only marginally, yet the performance impact is substantial (3.9 points). This suggests that the performance degradation from balanced sampling is not primarily about skewing the class distribution. Rather, it may stem from the sampler constraining which combinations of studies appear together in each batch, reducing the diversity of negative pairs available to the InfoNCE objective.

Overfitting is a recurring theme across experiments. Every model except the reproduction baseline shows a substantial gap between training and validation loss. The severity correlates with the degree of batch constraint: strict 50:50 case-level balancing shows the most extreme overfitting (training loss 0.002 vs.\ validation loss 1.8), followed by the section-level balanced models, then 100\% NAB with random sampling.

Small batch size likely exacerbates this problem. With a batch size of 8, each InfoNCE computation uses only 7 negatives per positive. Constraining the composition of these 8 samples further reduces the variety of negative pairs. For a batch size of 8 at 50:50 ratio, 4 normal and 4 abnormal studies produce at most 16 cross-class negative pairs per batch. Random sampling on the same data allows occasional batches of 7 abnormal and 1 normal, or 6 normal and 2 abnormal, creating a wider range of difficulty levels across batches. This stochastic variation may act as an implicit regularizer, analogous to how dropout introduces noise during training.

The overfitting patterns also suggest that validation loss is an unreliable proxy for downstream zero-shot performance. For 100\% NAB, the best zero-shot F1 (71.88\% at epoch 25) occurs well before the best validation loss epoch (epoch 50). This mismatch has practical implications: checkpoint selection based on validation loss alone may yield suboptimal models for zero-shot transfer.

The per-finding analysis reveals that findings differ substantially in their sensitivity to training conditions. Appendicitis and fracture are the most vulnerable to data reduction, dropping by 35 and 32 points respectively from 100\% NAB to 20\% NAB. Both conditions have relatively low prevalence, variable imaging presentations, and clinical descriptions that overlap with other pathologies. Appendicitis, for instance, may present as fat stranding, bowel wall thickening, or fluid collection, features shared with other acute abdominal conditions.

Conversely, pancreatic atrophy, hiatal hernia, and bowel obstruction show minimal sensitivity to data volume or batch composition. These findings have distinctive anatomical signatures, such as organ size reduction, herniation through the diaphragmatic hiatus, or dilated bowel loops with transition points, that create strong and consistent image-text correspondences.

Some findings respond differently to balanced sampling than to data reduction. Atherosclerosis drops from 81.67\% (100\% NAB) to 58.41\% (100\% NAB, 50:50) under balanced sampling on the same data, a 23-point decline, but only to 67.18\% under 5$\times$ data reduction (20\% NAB). This suggests that for certain findings, the diversity of batch compositions matters more than the total volume of training data. Atherosclerosis is typically described across multiple vascular regions, and restricting batch composition may limit the variety of vascular contexts the model encounters during training.

\subsection{Limitations}

Several limitations should be considered when interpreting these results. First, the batch composition and data scaling experiments were not all trained to convergence. The balanced-sampling models were stopped between epochs 20 and 36 due to overfitting, while the reproduction baseline ran to epoch 187. It is possible that modified learning rate schedules, stronger regularization, or early stopping based on downstream metrics rather than validation loss could improve the balanced-sampling results.

Second, the batch size of 8 is small relative to typical contrastive learning setups. SimCLR~\cite{chen2020simclr} and CLIP~\cite{radford2021clip} use batch sizes of 4,096 or more, where the effect of individual sample composition on the overall negative distribution is diluted. The original Merlin authors used gradient checkpointing for both visual and text encoders and trained with FP16 mixed precision, which enabled a maximum batch size of 18 on a single 48~GB A6000 GPU~\cite{blankemeier2024merlin}. We use the same hardware but found that gradient checkpointing incurred prohibitive overhead, exceeding two hours per epoch, which was impractical given the number of experiments in this study. Disabling gradient checkpointing reduced epoch time substantially but limited the batch size to 8. The sensitivity to batch composition observed in our experiments may be partly an artifact of this smaller batch size, and the findings may not generalize to the batch size of 18 used in the original work or to larger setups.

Third, the two experiment families use different datasets and samplers, which limits direct comparison. The batch composition experiments use the full 15,314-study dataset with section-level balancing, while the data scaling ablations use the 4,362-study NAB subset with case-level balancing. Differences in data quality, annotation consistency, and text characteristics between these subsets may confound cross-family comparisons.

Finally, we evaluate only zero-shot classification as the downstream task. Performance on other downstream tasks such as image retrieval, report generation, or few-shot learning might respond differently to batch composition and data scaling. The conclusions drawn here are specific to the zero-shot binary classification setting.

% ══════════════════════════════════════════════════════════════
\section{Conclusion}
\label{sec:conclusion}
% ══════════════════════════════════════════════════════════════

We reproduced the Merlin 3D vision-language model for abdominal CT, achieving a zero-shot macro F1 of 74.45\% across 30 findings (original: 73\%), and used this reproduction as a baseline to investigate batch composition and data scaling. Explicit control of the normal-to-abnormal ratio within training batches, whether via section-level or case-level balancing, consistently degrades zero-shot performance by 2.4 to 3.9 points relative to the original alternating batching strategy, which imposes no class constraints. Data scaling ablations show a sub-linear relationship between training set size and downstream F1, with individual findings varying substantially in data sensitivity. These results suggest that the stochastic diversity of random sampling serves as a more effective regularizer than engineered class ratios at the small batch sizes necessitated by 3D medical volumes, and that practitioners should assess whether their target pathologies fall among the data-hungry or data-efficient categories. Future work should examine whether these findings generalize to larger batch sizes, alternative contrastive objectives, and finer-grained or adaptive sampling strategies.

% ══════════════════════════════════════════════════════════════
% References
% ══════════════════════════════════════════════════════════════

\end{document}